\documentclass[journal]{IEEEtaes}

\usepackage[utf8]{inputenc}
\usepackage[T1]{fontenc}
\usepackage{amsmath,amsfonts,amssymb}
\usepackage{array}
\usepackage{booktabs}
\usepackage[caption=false,font=normalsize,labelfont=sf,textfont=sf]{subfig}
\usepackage{url}
\usepackage{graphicx}
\usepackage{orcidlink}  

\jvol{XX}
\jnum{XX}
\jmonth{XXXXX}
\paper{1234567}
\pubyear{2026}
\doiinfo{TAES.2026.Doi Number}

\usepackage{xcolor}
\newif\ifhighlight
\ifdefined\HAKFhighlight\highlighttrue\else\highlightfalse\fi
\definecolor{revblue}{RGB}{0,72,204}
\newcommand{\revtag}[1]{\ifhighlight\leavevmode{\color{revblue}\textsf{\scriptsize[#1]}}\ \fi}
\newcommand{\rev}[2][]{\ifhighlight\leavevmode{\color{revblue}\if\relax\detokenize{#1}\relax\else\revtag{#1}\fi#2}\else#2\fi}

\newtheorem{theorem}{Theorem}
\newtheorem{definition}{Definition}
\newtheorem{proposition}{Proposition}
\hypersetup{
  colorlinks=true,
  linkcolor=blue,   
  citecolor=blue,
  urlcolor=black,    
}

\makeatletter
\def\thebibliography#1{%
    \section*{\refname}%
    \addcontentsline{toc}{section}{\refname}%
    \normalfont\footnotesize
    \list{\@biblabel{\@arabic\c@enumiv}}%
    {%
    \settowidth\labelwidth{\@biblabel{#1}}%
    \leftmargin\labelwidth
    \advance\leftmargin\labelsep
    \itemindent0pt
    \itemsep0pt plus .5pt
    \usecounter{enumiv}%
    \let\p@enumiv\@empty
    \renewcommand\theenumiv{\@arabic\c@enumiv}}%
    \sloppy\clubpenalty4000\widowpenalty4000%
    \sfcode`\.=1000\relax}

\makeatother

\begin{document}

\title{Hybrid Adaptive Kalman Filtering for Data-Efficient Joint Tracking and Classification}

\author{JIHO LEE\,\orcidlink{0009-0003-9471-6893}}   
\member{Student Member, IEEE}               
\affil{Charles Stark Draper Laboratory, Inc., Cambridge, MA USA, and University of Colorado Boulder, Boulder, CO, USA}

\author{NISAR R. AHMED\,\orcidlink{0000-0002-7555-5671}}  
\member{Member, IEEE}     
\affil{University of Colorado Boulder, Boulder, CO, USA}

\author{REBECCA RUSSELL\,\orcidlink{0000-0002-4012-4513}} 
\affil{Charles Stark Draper Laboratory, Inc., Cambridge, MA, USA}

\receiveddate{Manuscript received XXXXX 00, 2026; revised XXXXX 00, 2026; accepted XXXXX 00, 2026. Date of publication XXXXX 00, 2026; date of current version XXXXX 00, 2026. \\ \\
DOI. NO. XX.XXXX/TAES.XX.XX \\\\
Refereeing of this contribution was handled by .\\\\ 
This work was supported by the Charles Stark Draper Laboratory, Inc., Cambridge, MA USA under the Draper Scholars Program.} 
\authoraddress{Authors’ addresses: Jiho Lee is with the Charles Stark Draper Laboratory, Inc., Cambridge, MA, USA, and the Ann and H. J. Smead Department of Aerospace Engineering Sciences, University of Colorado Boulder, Boulder, CO, USA (e-mail: jiho.lee@colorado.edu). Nisar Ahmed is with the Ann and H. J. Smead Department of Aerospace Engineering Sciences, University of Colorado Boulder, Boulder, CO, USA (e-mail: nisar.ahmed@colorado.edu). Rebecca Russell is with the Charles Stark Draper Laboratory, Inc., Cambridge, MA, USA (e-mail: rrussell@draper.com). (\textit{Corresponding author: Jiho Lee}.)}

\markboth{LEE ET AL.}{HYBRID ADAPTIVE KALMAN FILTERING FOR DATA-EFFICIENT JOINT TRACKING AND CLASSIFICATION}
\maketitle

\begin{abstract}Kalman filtering performance is highly sensitive to model mismatch and noise covariance tuning. Classical adaptive filtering methods address these issues by estimating model parameters or noise covariances from measurement data but rely on a predefined model structure. Learning-based methods provide greater flexibility but typically require large supervised datasets and often lack well-calibrated uncertainty estimates. This article proposes a self-supervised Hybrid Adaptive Kalman Filter (HAKF) that learns time-varying corrections to the system dynamics and process-noise covariance directly from measurements while jointly estimating the measurement-noise covariance. Because these corrections are learned from measurements alone, we analyze the structural identifiability of the corresponding time-invariant parameterizations. For a constant-velocity model with position-only measurements, an unrestricted process-noise covariance is not identifiable even when the dynamics are known, whereas a physically structured parameterization that couples the dynamics and process covariance is jointly identifiable with the measurement-noise covariance. This analysis motivates a structured HAKF variant, compared with a flexible variant that adapts the dynamics and process covariance independently. Because HAKF preserves the probabilistic structure of the Kalman filter, the innovation likelihood also enables model classification through generalized Bayesian inference. In Isaac Lab simulations, the structured HAKF improves tracking accuracy over a fitted Kalman filter across all training-set sizes. When training data are limited, its advantage over the flexible variant is largest, and it outperforms a supervised recurrent neural network in both tracking and classification. On recorded DPJAIT flight trajectories, it achieves the best tracking and classification performance with near-consistent state covariance, whereas the flexible variant becomes overconfident.
\end{abstract}

\section*{Nomenclature}
\begin{IEEEdescription}[\IEEEusemathlabelsep\IEEEsetlabelwidth{$\Delta F_k^{\mathrm{ML}}$}]
\item[\textit{Symbol}] \textit{Definition}
\item[$x_k$] State vector at time step $k$.
\item[$z_k$] Measurement vector at time step $k$.
\item[$F_k$] State transition matrix.
\item[$F^{\mathrm{CV}}$] Nominal constant-velocity state transition matrix.
\item[$F_k^{\mathrm{ML}}$] Learned state transition matrix.
\item[$\Gamma_k$] Velocity-dynamics parameters of the structured variant.
\item[$\Sigma_k$] Acceleration-noise spectral density of the structured variant.
\item[$\Delta t$] Sampling interval.
\item[$H_k$] Observation matrix.
\item[$Q_k$] Process noise covariance matrix.
\item[$Q_k^{\mathrm{ML}}$] Learned process noise covariance matrix.
\item[$R_k$] Measurement noise covariance matrix.
\item[$R^{\mathrm{ML}}$] Learned time-invariant measurement noise covariance matrix.
\item[$w_k$] Process noise, $w_k \sim \mathcal{N}(0,Q_k)$.
\item[$v_k$] Measurement noise, $v_k \sim \mathcal{N}(0,R_k)$.
\item[$\hat{x}_{k|k-1}$] Predicted state estimate.
\item[$P_{k|k-1}$] Predicted state covariance.
\item[$\hat{x}_{k|k}$] Posterior state estimate.
\item[$P_{k|k}$] Posterior state covariance.
\item[$\nu_k$] Innovation (measurement residual).
\item[$S_k$] Innovation covariance.
\item[$K_k$] Kalman gain.
\item[$e_{x,k}$] State estimation error.
\item[$\phi_k$] Feature vector used by the adaptive neural network.
\item[$f_\theta(\cdot)$] Neural network parameterized by $\theta$.
\item[$\lambda$] Interpolation hyperparameter between nominal and learned dynamics.
\item[$W$] Length of the innovation-statistics window.
\item[$\mathcal{L}(\theta)$] Training loss function.
\item[$M_j$] Candidate model or class $j$.
\item[$\alpha$] Generalized Bayesian inference tempering parameter.
\item[$\epsilon_{x,k}$] Normalized estimation error squared.
\item[$\epsilon_{z,k}$] Normalized innovation squared.
\end{IEEEdescription}

\section{Introduction}
\ifhighlight\noindent{\color{revblue}\textsf{\scriptsize Tracked-changes version: the major revisions made in response to the review comments are shown in blue, and the bracketed labels give the comments they address.}}\par\fi

Kalman filters (KFs) and their variants have been widely used for tracking, navigation, and state estimation because of their principled probabilistic framework and computational efficiency \cite{kalman_new_1961, bar-shalom_estimation_2001, simon_optimal_2006}. In practice, however, the performance of a KF depends heavily on accurate system models and carefully tuned noise covariances. Model mismatch, unmodeled disturbances, and mistuned noise statistics can lead to biased state estimates, inconsistent uncertainty quantification, and degraded downstream inference. Obtaining reliable model parameters often requires expert knowledge, system identification, and manual tuning.

Classical adaptive filtering methods, extensively studied since the 1970s, address these challenges by adapting filter parameters directly from measurement data \cite{mehra_approaches_1972}. Techniques such as covariance matching, correlation analysis, and likelihood-based estimation enable parameter adaptation without requiring ground-truth state information \cite{zhang_identification_2020}, providing an important foundation for measurement-based filter tuning. However, they typically assume a predefined parametric state-space structure. As a result, structural mismatch that falls outside the assumed model can induce biases and temporally correlated innovations that cannot be corrected through parameter adaptation alone.
 
Recent advances in machine learning have motivated a growing body of work that integrates neural networks with classical filtering frameworks \cite{jin_new_2021}. These learning-augmented approaches aim to improve estimation performance by learning system dynamics, filter parameters, or correction terms directly from data. While promising, existing learning-based filtering methods typically rely on supervised training with access to ground-truth states and are optimized using accuracy-based loss functions. As a result, these methods primarily improve point-estimation accuracy but often neglect the statistical consistency of the estimated uncertainty, which is critical for assessing filter reliability. Furthermore, many learning-based approaches assume access to large-scale training datasets, often generated in simplified simulation environments, which may not be available in real-world systems. 

\begin{figure}[t]
    \centering
    \includegraphics[width=\linewidth]{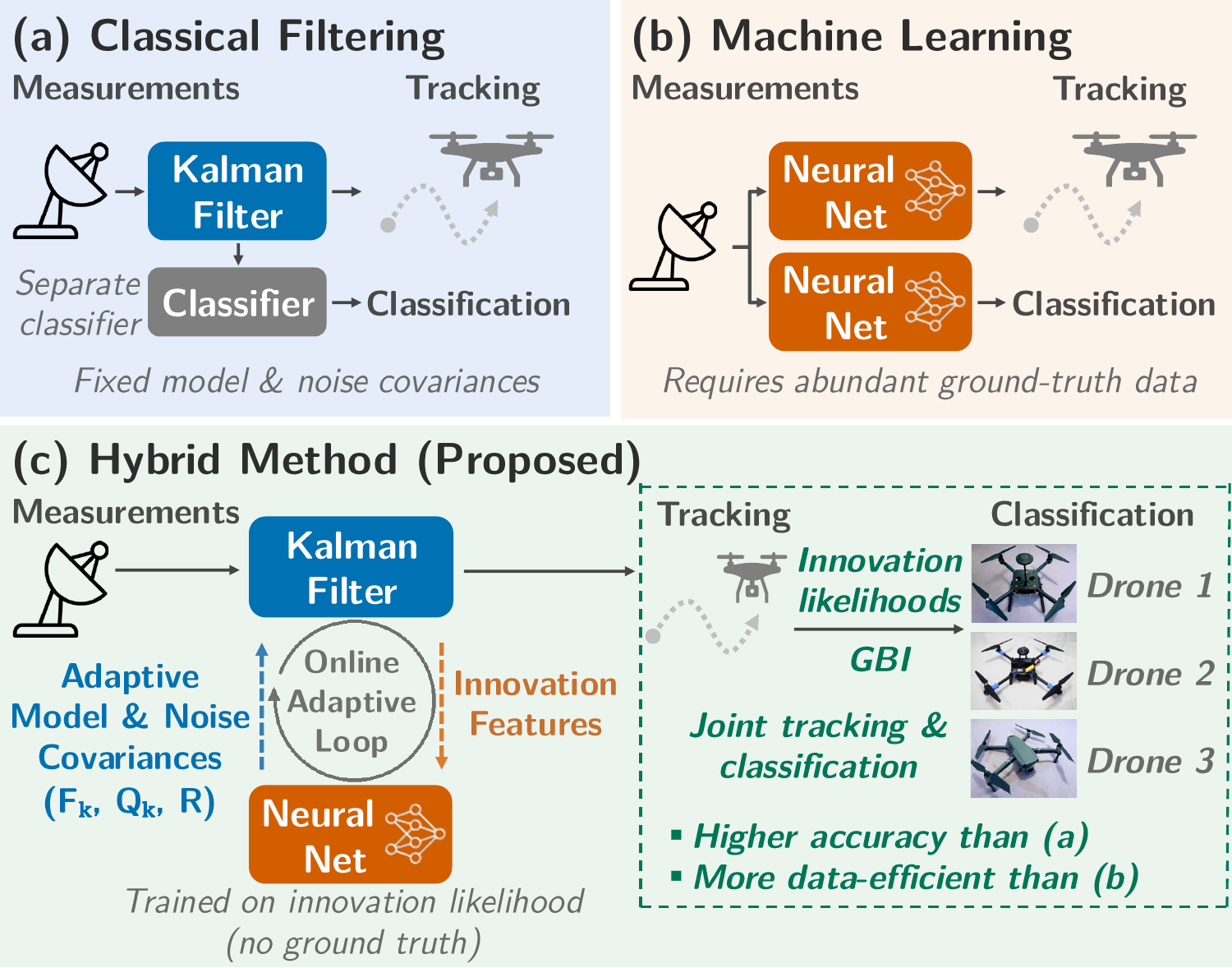}
    \caption{Comparison between classical filtering, purely learning-based methods, and the proposed hybrid framework. (a) Classical KF relies on a fixed model and noise covariances tuned with domain expertise, with classification performed using a separate classifier. (b) Purely learning-based approaches use neural networks to directly predict system states and classes, typically requiring abundant ground-truth training data. (c) The proposed Hybrid Adaptive Kalman Filter combines model-based filtering with data-driven adaptation trained on the innovation likelihood without ground-truth states, enabling accurate state estimation and probabilistic model classification through generalized Bayesian inference under limited-data conditions.}
    \label{fig:figure_1}
\end{figure}

These limitations pose challenges for many safety-critical systems where ground-truth state information may be unavailable, uncertainty calibration is essential, and collecting large labeled datasets can be costly or infeasible. Motivated by these limitations, this article proposes a Hybrid Adaptive Kalman Filter (HAKF) that combines model-based filtering with data-driven learning. The proposed framework augments a classical KF with a neural network that learns time-varying corrections to the nominal dynamics and process-noise covariance, while a constant measurement-noise covariance is learned jointly from the same measurement data. By incorporating structural inductive bias from the underlying physics-based model and restricting learning to components most susceptible to mismatch, the proposed approach enables data-efficient adaptation while preserving the probabilistic structure of the filter. 

In addition to improving state estimation, the measurement likelihood produced by the hybrid filter naturally enables model classification through a generalized Bayesian inference (GBI) framework \cite{knoblauch_generalized_2019}. This allows the same filtering architecture to perform both probabilistic state estimation and model classification. Fig.~\ref{fig:figure_1} illustrates the overall conceptual differences between classical KF, purely learning-based approaches, and the proposed hybrid framework.

\rev[R1\_3, R2\_1]{Learning the corrections from measurements alone, however, raises the question of which parameters can be uniquely determined from the data. Distinct combinations of dynamics and noise covariances may produce identical measurement statistics, in which case they cannot be distinguished regardless of the amount of data \cite{zhang_identification_2020}. In this paper, we study a three-dimensional position--velocity tracking problem using a nominal constant-velocity (CV) model, which serves as a tractable baseline rather than an assumption that the observed drone trajectories have constant velocity. For the time-invariant CV model with position-only measurements, we show that an unrestricted process-noise covariance is not identifiable even when the dynamics are known, whereas a physically structured parameterization that couples the dynamics and process-noise covariance is identifiable together with the measurement-noise covariance. This analysis motivates a structured HAKF variant, which we compare with a flexible variant that adapts the dynamics and process-noise covariance independently.}

We evaluate the proposed hybrid method on both recorded drone trajectories from the multimodal indoor 3D drone tracking dataset (DPJAIT) \cite{rosner_multimodal_2025} and high-fidelity simulated trajectories generated using Isaac Lab \cite{nvidia_isaac_2025}. In the Isaac Lab simulations, the structured HAKF improves tracking accuracy over a fitted classical KF across all tested training-set sizes, with the largest advantage over the flexible variant when training data are limited. In this low-data regime, both HAKF variants also outperform a supervised recurrent neural network in tracking and classification, whereas the supervised network becomes more accurate as larger training datasets become available. On the DPJAIT trajectories, the structured HAKF achieves the best tracking and classification performance among the compared methods with near-consistent state covariance.

To summarize, the key contributions of this article are as follows.
\begin{itemize}
    \item[1)] A self-supervised HAKF framework inspired by classical adaptive KF that integrates model-based filtering with neural network-based adaptive model correction without requiring ground-truth state supervision.
    \item[2)]  \rev[R1\_3, R2\_1]{A structural identifiability analysis of the time-invariant CV parameterizations, showing that an unrestricted process-noise covariance is not identifiable from position-only measurements, whereas the physically structured parameterization is. This guides the design of the structured HAKF variant.}
    \item[3)] A GBI framework that enables probabilistic model classification using the innovation likelihood produced by the hybrid filter.
    \item[4)] Experimental validation on both real-world and simulated datasets demonstrating data-efficient joint tracking and classification.
\end{itemize}

The rest of this article is organized as follows. Section \ref{sec:Related_work} reviews related work, and Section \ref{sec:HAKF_section} presents the problem formulation, the CV parameterizations and their identifiability, and the proposed HAKF framework. Section \ref{sec:experimental_setup} describes the datasets, baselines, and evaluation metrics, and Section \ref{sec:results} reports experimental results for both state estimation and model classification. Finally, Section \ref{sec:conclusion} concludes the article and discusses limitations and directions for future work.

\section{Related Work}
\label{sec:Related_work}
\subsection{Kalman Filter (KF)} 
The KF provides a Bayesian optimal estimator for linear dynamical systems subject to Gaussian noise \cite{kalman_new_1961}. Consider the following discrete-time state-space model:
\begin{align}
\label{dt-system-1}
x_k &= F_k x_{k-1} + w_k, \\
z_k &= H_k x_k + v_k.
\label{dt-system-2}
\end{align}
The Nomenclature summarizes the KF notation used throughout the article.

If the linear-Gaussian assumptions of the KF are satisfied and the system model is correctly specified, then the following dynamical consistency conditions must hold \cite{bar-shalom_estimation_2001}: \\
\textbf{1) Unbiased}
\begin{align}
\label{eq:kf_consistency1}
\mathbb{E}[e_{x,k}] = 0  \hspace{3.mm} \forall k.
\end{align}
\textbf{2) Consistency}
\begin{align}
\label{eq:innov}
\mathbb{E}[e_{x,k}e_{x,k}^\top] = P_{k|k}  \hspace{3.mm} \forall k.
\end{align}
\textbf{3) Innovation Whiteness}
\begin{align}
\nu_{k} &\sim \mathcal{N}(0,S_{k}), \\
\mathbb{E}[\nu_{k}] &= 0, \\
\mathbb{E}[\nu_{k}\nu_{j}^\top]
&=
\delta_{kj}S_{k}.
\label{eq:kf_consistency2}
\end{align}
In practice, the assumed linear-Gaussian system model is often incorrect due to modeling errors and mistuned noise covariances. Such mismatch leads to biased estimation errors and temporally correlated innovation sequences \cite{zhang_identification_2020}, violating the dynamical consistency conditions \eqref{eq:kf_consistency1}--\eqref{eq:kf_consistency2} required for optimal Bayesian inference. Hence, violations of these properties provide informative signals of model mismatch that can be exploited for adaptive model correction.

\subsection{Adaptive Kalman Filter}

To address this issue, adaptive Kalman filtering methods have been extensively studied since the 1970s \cite{mehra_approaches_1972}. Existing approaches can be broadly categorized into Bayesian inference, maximum likelihood estimation, covariance-matching, and correlation methods. More recently, consistency-based methods have been proposed that tune filter noise covariances by enforcing their statistical consistency.

Bayesian methods treat unknown model parameters as latent random variables and estimate them jointly with the system state through recursive inference. The posterior distribution of the parameters is updated sequentially using Bayes' rule; however, exact inference is generally intractable and typically requires approximate techniques like Monte Carlo sampling, which can introduce computational overhead and limit real-time application \cite{hilborn_optimal_1969,matisko_noise_2013,mehra_identification_1970}. \rev[R1\_1]{Variational Bayesian adaptive KFs avoid sampling by recursively approximating the joint posterior of the state and the noise covariances \cite{sarkka_recursive_2009,huang_novel_2018}. }

Maximum likelihood methods estimate unknown model parameters by maximizing the likelihood of the observed measurement sequence under an assumed probabilistic model \cite{dempster_maximum_1977,shumway_approach_1982}. In the marginal likelihood formulation, the state variables are treated as latent and integrated out, and parameters are obtained by solving
\begin{align}
\label{marginal_likelihood}
\theta^* = \arg\max_\theta \, p_\theta(z_{1:T}),
\end{align}
where $\theta$ denotes the set of unknown model parameters and \rev[R1\_5]{$T$ is the number of measurements in the sequence}. These methods typically require iterative optimization of a non-convex likelihood objective, making them sensitive to initialization and susceptible to local optima convergence \cite{zhang_identification_2020}. 

Covariance-matching techniques enforce consistency between the empirical covariance of the innovation sequence and its theoretical covariance predicted by the KF. More formally,
\begin{align}
\frac{1}{m}\sum_{i=k-m+1}^{k} \nu_i \nu_i^\top 
\;\approx\;
S_{k},
\end{align}
where $m$ is an empirically chosen finite window length used for statistical smoothing \cite{mehra_approaches_1972}.

Correlation-based methods exploit the fact that, under correct model specification, the innovation sequence should be temporally uncorrelated. Nonzero innovation autocorrelation provides a direct signal of misspecification, which can be used to adapt filter parameters \cite{mehra_approaches_1972,zhang_identification_2020}.

Lastly, recent work proposes consistency-based methods that tune the filter noise covariances such that the state estimation errors and innovations are statistically consistent \cite{chen_weak_2018,chen_kalman_2024}. The normalized estimation error squared (NEES) and normalized innovation squared (NIS) statistics, 
\begin{align}
\label{eq:nees}
\epsilon_{x,k} &= e_{x,k}^\top P_{k|k}^{-1} e_{x,k}, \\
\epsilon_{z,k} &= \nu_k^\top S_{k}^{-1} \nu_k,
\label{eq:nis}
\end{align}
follow chi-squared distributions under correct model specification \cite{bar-shalom_estimation_2001}. These methods adjust the noise parameters by matching the empirical statistics to their corresponding theoretical moments.

\rev[R1\_1]{A related question is which parameters can be recovered from measurements at all. Mehra's correlation-based identification method assumes a known structure for the process-noise covariance and requires the number of unknown elements in $Q$ to be less than the product of the state and measurement dimensions \cite{mehra_identification_1970}. More recently, Zhang et al. established rank-based conditions for identifying the process and measurement noise covariances \cite{zhang_identification_2020}. However, both analyses assume known system dynamics. When the dynamics and noise covariances are learned jointly, as in HAKF, additional ambiguities can arise, which we analyze for the CV model in Section~\ref{sec:identifiability_cv}.}

Our proposed HAKF architecture integrates principles from classical adaptive filtering methods, including maximum likelihood, covariance matching, and correlation approaches, to enable adaptation without requiring ground-truth states. Accordingly, NEES and NIS are used in the experiments to evaluate the consistency of the resulting estimates.

\subsection{Hybrid Kalman Filter}
Recent work combines KF with learned neural components to mitigate model mismatch and unknown noise statistics. These approaches are commonly referred to as hybrid methods and can be categorized into the following three categories: state-correction, model-learning, and gain-learning methods. 

State-correction approaches augment the Kalman filtering framework with a learned neural component that compensates for estimation errors. In these methods, the neural network typically learns a residual mapping to correct either the predicted state or the innovation term without modifying the assumed system model \cite{jin_new_2021,shaukat_multi-sensor_2021,nguyen_calibration_2015}. While such approaches can improve point-estimation accuracy, they rely on post hoc compensation and do not explicitly enforce statistical consistency of the associated estimates.

Model-learning methods learn unknown system parameters directly from data. In these approaches, neural networks parameterize components of the underlying state-space model, enabling the representation of complex nonlinear dynamics beyond classical linear-Gaussian assumptions \cite{fraccaro_disentangled_2017,karl_deep_2017,krishnan_deep_2015}. This improves modeling flexibility but often replaces explicitly structured model components with learned parameterizations, making the resulting filter harder to interpret in classical state-space terms.

Lastly, gain-learning methods retain the prediction step of the KF while replacing the analytically derived measurement update step with a learned neural update operator. Representative approaches such as KalmanNet and its variants explicitly replace or augment the analytical Kalman gain computation with recurrent neural networks \cite{revach_kalmannet_2022,dahan_bayesian_2025,ko_cholesky-kalmannet_2025,mortada_recursive_2025}. These methods have been shown to improve state estimation accuracy under partially known dynamics without requiring explicit identification of the process or measurement noise statistics. \rev[R1\_1]{Unsupervised variants train the network from observation-prediction errors without ground-truth states \cite{revach_unsupervised_2022}, but they still learn the Kalman gain rather than explicitly parameterizing the dynamics and noise covariances.}

In contrast, our proposed HAKF preserves the analytical Kalman prediction and update recursion and uses learning only to adapt time-varying corrections to the dynamics and process noise while jointly estimating the measurement-noise covariance. \rev[R1\_1]{Like unsupervised KalmanNet, HAKF can be trained directly from measurements without ground-truth states. However, HAKF explicitly parameterizes the system dynamics and noise covariances rather than learning the Kalman gain.} This distinction is important because while KF is often interpreted as a minimum mean-square error (MMSE) estimator, it fundamentally performs recursive Bayesian inference by propagating both the posterior mean and covariance. Consequently, training and evaluating hybrid filters solely in terms of point-estimation accuracy disregards the uncertainty information captured by the filter that is essential for assessing filter reliability. Moreover, existing hybrid methods typically assume abundant training data, which are often generated in simplified simulation environments. \rev[R1\_4]{For these reasons, this article does not aim to benchmark HAKF against all existing hybrid filters. Instead, the experiments compare HAKF with a classical KF fitted with the same objective, a flexible HAKF variant, a supervised RNN, and classifiers trained on labeled measurement segments, each of which isolates one design choice of HAKF (Section~\ref{sec:baselines}).} 

\section{HAKF: Hybrid Adaptive Kalman Filter}
\label{sec:HAKF_section}
In this section, we present a self-supervised HAKF that learns corrections to the system dynamics and estimates process and measurement noise covariances while preserving the analytic Bayesian filtering structure. We first describe the problem formulation and the CV model used in this study, then examine parameter identifiability before presenting the proposed hybrid filtering architecture, training objective, and classification method.

\subsection{Problem Formulation}
In practice, the discrete-time linear dynamical system described in \eqref{dt-system-1}--\eqref{dt-system-2} is often misspecified due to unmodeled dynamics and inaccurate noise statistics. The objective of the proposed HAKF is therefore to adaptively learn corrections to both the nominal system dynamics and process-noise covariances, while jointly estimating the measurement-noise covariance from observed measurement sequences. 

Specifically, a neural network parameterized by
$\theta$ maps a feature vector $\phi_k$ to
time-varying model parameters,
\begin{align}
\eta_k = f_\theta(\phi_k),
\end{align}
which determine the transition matrix
$F_k^{\mathrm{ML}}$ and process-noise covariance
$Q_k^{\mathrm{ML}}$.
The resulting state evolution and observation models are
\begin{align}
x_k &= F_k^{\mathrm{ML}}x_{k-1}+\tilde{w}_k,
\qquad
\tilde{w}_k\sim\mathcal{N}(0,Q_k^{\mathrm{ML}}),\\
z_k &= H_kx_k+v_k,
\qquad
v_k\sim\mathcal{N}(0,R^{\mathrm{ML}}).
\end{align}
The observation matrix $H_k$ is known, while the time-invariant measurement-noise covariance $R^{\mathrm{ML}}$ is learned jointly with the model parameters rather than predicted at each time step. \rev[R2\_4]{As in the standard KF and recent hybrid filters~\cite{dahan_bayesian_2025,mortada_recursive_2025}, the process and measurement noises are assumed to be mutually independent. In both datasets, the measurement noise is generated independently of the motion.}

The mapping from $\eta_k$ to $(F_k^{\mathrm{ML}},Q_k^{\mathrm{ML}})$
depends on the parameterization. The flexible variant independently predicts the velocity-transition block $\Delta F_k^{\mathrm{ML}}$ and
$Q_k^{\mathrm{ML}}$, yielding $(F_k^{\mathrm{flex}},Q_k^{\mathrm{flex}})$. In contrast, the structured variant predicts physical parameters $(\Gamma_k,\Sigma_k)$,
from which $(F_k^{\mathrm{str}},Q_k^{\mathrm{str}})$ are jointly derived. These parameterizations are detailed in Section~\ref{sec:cv_parameterization}.

\subsection{CV Motion Model and Adaptive Parameterization}
\label{sec:cv_parameterization}
The HAKF framework is not restricted to a particular dynamical model. In this work, we consider three-dimensional position--velocity tracking under a nominal CV model. With sampling interval $\Delta t>0$, the state and position-only measurement model are
\begin{align}
x_k &= [p_{x,k},p_{y,k},p_{z,k},\dot p_{x,k},\dot p_{y,k},\dot p_{z,k}]^\top, \\
 F^{\mathrm{CV}} &=
 \begin{bmatrix}I_3 & \Delta t I_3\\0&I_3\end{bmatrix},
 \qquad H=\begin{bmatrix}I_3&0\end{bmatrix}.
 \label{eq:cv_model_matrices}
\end{align}
The CV model serves as a nominal kinematic approximation rather than assuming that the tracked objects move at constant velocity. Unmodeled accelerations and other forces therefore motivate adaptive corrections to the velocity dynamics and process-noise covariance.

A fundamental challenge in measurement-based adaptive filtering is parameter identifiability: distinct combinations of system dynamics and noise covariances may produce the same measurement statistics, making their unique recovery impossible. To investigate how physically motivated constraints can reduce these ambiguities, we consider two HAKF parameterizations: flexible and structured.

In the flexible parameterization, the nominal CV position update is preserved, while the velocity-transition block and process-noise covariance are parameterized separately:
\begin{align}
F_k^{\mathrm{flex}} &=
\begin{bmatrix}
I_3 & \Delta t I_3\\
0 & (1-\lambda)I_3+\lambda\Delta F_k^{\mathrm{ML}}
\end{bmatrix},\\
Q_k^{\mathrm{flex}} &= L_kD_kL_k^\top,
\end{align}
where $\lambda\in[0,1]$ is an interpolation hyperparameter that controls the contribution of the fully parameterized $\Delta F_k^{\mathrm{ML}}\in\mathbb{R}^{3\times3}$, and the network outputs a unit lower-triangular $L_k$ and positive diagonal $D_k$ to ensure $Q_k^{\mathrm{flex}}\succ0$~\cite{pinheiro_unconstrained_1996}. No physical coupling is imposed between $F_k^{\mathrm{flex}}$ and $Q_k^{\mathrm{flex}}$.

\rev[R1\_3, R2\_1]{In contrast, the structured parameterization couples the transition matrix and process-noise covariance through the continuous-time stochastic dynamics}

\begin{align}
\label{eq:structured_cv_ct}
\dot{x}(t) &= A(\Gamma_k)x(t)+Gw_a(t),\\
A(\Gamma_k) &=
\begin{bmatrix}
0 & I_3\\
0 & -\Gamma_k
\end{bmatrix},
\qquad
G=\begin{bmatrix}0\\I_3\end{bmatrix},
\end{align}
where $\Gamma_k\in\mathbb{R}^{3\times3}$ is a diagonal matrix of signed velocity-dynamics parameters, with positive values damping and negative values amplifying the velocity, and $w_a(t)$ is zero-mean white acceleration noise with diagonal spectral-density matrix $\Sigma_k\succeq0$. Assuming these parameters remain constant within each sampling interval, exact discretization yields
\begin{align}
F_k^{\mathrm{str}}
&= \exp\bigl(A(\Gamma_k)\Delta t\bigr),\\
Q_k^{\mathrm{str}}
&= \int_0^{\Delta t}
e^{A(\Gamma_k)\tau}G\Sigma_k G^\top
e^{A(\Gamma_k)^\top\tau}\,d\tau.
\label{eq:structured_cv_discretization}
\end{align}
Unlike the flexible parameterization, the structured model derives both $F_k^{\mathrm{str}}$ and $Q_k^{\mathrm{str}}$ from shared physical parameters,
thereby restricting the admissible model family. The nominal CV transition is recovered when $\Gamma_k=0$.

In both HAKF variants, the diagonal, time-invariant measurement-noise covariance $R^{\mathrm{ML}}$ is learned jointly with the adaptive model parameters. The identifiability analysis in the following subsection considers the corresponding time-invariant models, with $\Gamma=\operatorname{diag}(\gamma_x,\gamma_y,\gamma_z)$, $\Sigma=\operatorname{diag}(\sigma_x^2,\sigma_y^2,\sigma_z^2)$, and $R=\operatorname{diag}(r_x,r_y,r_z)$, where all parameters are constant over time.

%%% Identifiability Theorem
\subsection{Identifiability of the CV parameterization}
\label{sec:identifiability_cv} 
\rev[AE, R1\_3, R2\_1]{We now examine the identifiability of the time-invariant counterparts of the CV parameterizations introduced in Section~\ref{sec:cv_parameterization}. Zhang et al. established rank-based conditions for identifying process and measurement noise covariances from transformed innovation statistics under known system dynamics \cite{zhang_identification_2020}. Here, we examine the additional ambiguity introduced by learning the dynamics and noise jointly.}

\begin{definition}[Structural Identifiability]
Let $\vartheta$ denote a constant model-parameter vector and $\mu$ an unknown Gaussian initial-state distribution. Define $\mathcal{S}(\vartheta,\mu)$ as the collection of all first- and second-order statistics of the observed measurement sequence $Z=\{z_k\}_{k\geq0}$. The parameterization is structurally identifiable if,
for any admissible $\vartheta$, $\vartheta'$, $\mu$, and $\mu'$,
\begin{align}
\mathcal{S}(\vartheta,\mu)
=
\mathcal{S}(\vartheta',\mu')
\implies
\vartheta=\vartheta'.
\end{align}
For the linear-Gaussian models considered here, the first- and second-order statistics fully characterize the joint measurement distribution.
\end{definition}

\begin{proposition}[Non-Identifiability of Unrestricted $Q$]
\label{thm:full_q_nonidentifiable}%
\rev[R2\_1]{Consider the time-invariant CV model with known $F=F^{\mathrm{CV}}$ from \eqref{eq:cv_model_matrices}, position-only $H=[I_3\ \ 0]$, and known $R$, but unknown $Q\in\mathbb{S}_{++}^{6}$ and unknown initial covariance $P_0\in\mathbb{S}_{++}^{6}$. Then $Q$ is not structurally identifiable from position-only measurements.}
\end{proposition}
\noindent\emph{Proof.} For any nonzero symmetric $\Omega\in\mathbb{S}^{3}$ sufficiently small to preserve the positive definiteness of $P_0'$ and $Q'$, define
\begin{align}
E &= \begin{bmatrix}0&0\\0&\Omega\end{bmatrix},\quad
P_0'=P_0+E,\quad
Q' = Q+E-FEF^\top.
\label{eq:full_q_null_family}
\end{align}
The state covariances $P_k=\operatorname{Cov}(x_k)$ satisfy $P_k'=P_k+E$ by induction. For any $j\geq k$, the difference in the measurement cross-covariances is
\begin{align}
\operatorname{Cov}(z_k',z_j')
-\operatorname{Cov}(z_k,z_j) =HE(F^{j-k})^\top H^\top=0,
\end{align}
since $HE=0$. Thus, both models have identical measurement covariances at all time steps. With the same initial mean, their Gaussian measurement sequences have identical joint distributions. However, $Q'\neq Q$ for $\Delta t>0$ and $\Omega\neq0$. Therefore, $\Omega$ defines a family of null directions along which the observable measurement statistics remain unchanged, establishing non-identifiability.

\rev[R2\_1]{For KFs initialized with $P_0$ and $P_0'$, the Kalman gains, state estimates, and innovations along this family are identical, while the predicted and posterior covariances differ only by $E$. Thus, the ambiguity affects only the velocity block of the estimated state covariance, not the state estimates. Because this ambiguity persists even when $F$ is known, allowing the velocity-transition block to vary cannot restore identifiability. Therefore, the time-invariant flexible parameterization is also not jointly identifiable in the velocity-transition block and $Q$. Appendix~\ref{app:likelihood_geometry} illustrates this non-identifiable covariance family as a likelihood ridge and contrasts it with the structured parameterization.}

\begin{theorem}[Identifiability of Structured CV]
\label{thm:structured_cv_identifiability}%
\rev[R1\_3, R2\_1]{Consider the one-dimensional structured CV model defined by \eqref{eq:structured_cv_ct}--\eqref{eq:structured_cv_discretization}, with known $\Delta t>0$, unknown constant $\gamma\in\mathbb{R}$, acceleration-noise intensity $\sigma^2>0$, and measurement-noise variance $r>0$. Assume that the Gaussian initial state, white Gaussian acceleration noise, and white Gaussian measurement noise are mutually independent, and the initial-state mean and covariance are unknown. Then $(\gamma,\sigma^2,r)$ is structurally identifiable from the position-measurement sequence. The same result holds for the three-axis model with diagonal $\Gamma$, $\Sigma$, and $R$.}
\end{theorem}
\noindent\emph{Proof.} Let $d_k=z_{k+1}-z_k$ and $C_j(k)=\operatorname{Cov}(d_k,d_{k+j})$. Writing $\rho=e^{-\gamma\Delta t}$, the lagged covariances satisfy $C_3(k)=\rho C_2(k)$, with $C_2(k)>0$. Thus,
\begin{align}
\rho=\frac{C_3(k)}{C_2(k)},\qquad
\gamma=-\frac{\log\rho}{\Delta t}.
\label{eq:identifiability_gamma_recovery}
\end{align}
With $u_k=d_k-\rho d_{k-1}$ and $\Lambda_j=\operatorname{Cov}(u_k,u_{k+j})$, the remaining parameters satisfy
\begin{align}
\Lambda_2&=\rho r,\nonumber\\
\Lambda_0&=\ell_0(\gamma)\sigma^2+\kappa_0(\rho)r,
\label{eq:identifiability_noise_recovery}
\end{align}
where $\ell_0(\gamma)>0$ and $\kappa_0(\rho)=1+(1+\rho)^2+\rho^2$. Hence $r=\Lambda_2/\rho$ and $\sigma^2=[\Lambda_0-\kappa_0(\rho)r]/\ell_0(\gamma)$ are uniquely determined. The derivation, positivity conditions, and extension to the three-dimensional model are given in Appendix~\ref{app:structured_identifiability}.

The proposition establishes the non-identifiability of unrestricted $Q$ even when $F$ and $R$ are known, whereas the theorem shows how restricting $Q$ to one physically structured intensity per axis enables joint identification of $F$ and $Q$ along with $R$. \rev[R2\_1]{However, both results concern the structural identifiability of linear time-invariant models and do not guarantee accurate parameter estimation from finite data or identifiability of arbitrary time-varying neural-network outputs.}

\subsection{Architecture}
\label{sec:Architecutre}
\begin{figure}[t]
    \centering
    \includegraphics[width=\linewidth]{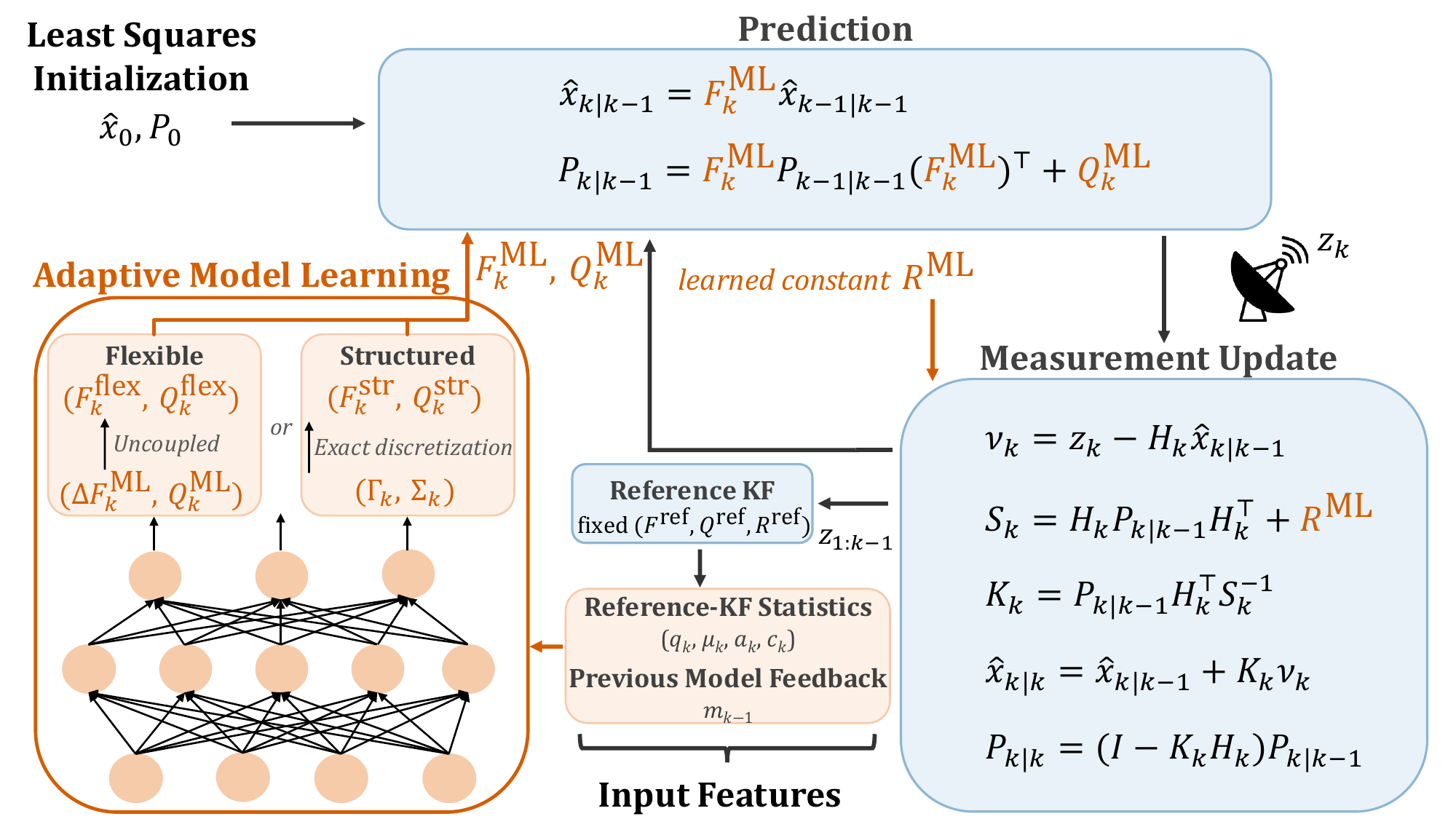}
    \caption{HAKF architecture. A fixed reference KF provides the reference-KF statistics $(q_k,\mu_k,a_k,c_k)$, which are combined with previous model feedback $m_{k-1}$. The network generates time-varying models using either the flexible or structured parameterization. $R^{\mathrm{ML}}$ is learned jointly and held constant over time.}
    \label{fig:architecture}
\end{figure}

Fig.~\ref{fig:architecture} illustrates the proposed HAKF architecture, which preserves the standard Kalman prediction and measurement correction recursion. The reference KF uses a fixed nominal transition model $(F^{\mathrm{ref}}=F^{\mathrm{CV}})$ while the time-invariant noise covariances $(Q^{\mathrm{ref}},R^{\mathrm{ref}})$ are fitted by minimizing the cost function defined in Section~\ref{sec:objective}.

Let $\nu_j^{\mathrm{ref}}$ and $S_j^{\mathrm{ref}}$ be the reference KF innovation and its covariance. For each position axis $i\in\{1,2,3\}$, define the normalized innovation and preceding reference
velocity estimate as
\begin{align}
e_{j,i}&=\frac{\nu_{j,i}^{\mathrm{ref}}}
{\sqrt{S_{j,ii}^{\mathrm{ref}}}},&
\xi_{j,i}&=
\bigl[\hat x_{j-1|j-1}^{\mathrm{ref}}\bigr]_{3+i}.
\label{eq:architecture_ref_features}
\end{align}
Thus, $\xi_{j,i}$ is the velocity estimate after the reference KF update at $j-1$, but before the measurement at $j$ is processed. This avoids using a state estimate corrected by the same measurement that produces $e_{j,i}$.

\rev[R1\_5]{Let $\mathcal{W}_k=\{k-W,\ldots,k-1\}$ be a window of length $W$, where $W$ is a hyperparameter, and let $\mathcal{W}_k^-=\{k-W+1,\ldots,k-1\}$}. For $k\le W$, the window is zero-padded. The per-axis innovation statistics are
\begin{align}
\label{eq:architecture_correlations}
q_{k,i}&=\log\left(\varepsilon_q+
\frac{1}{W}\sum_{j\in\mathcal{W}_k}e_{j,i}^2\right),
\nonumber\\
\mu_{k,i}&=\frac{1}{W}
\sum_{j\in\mathcal{W}_k}e_{j,i},
\nonumber\\
a_{k,i}&=
\frac{\sum_{j\in\mathcal{W}_k^-}e_{j,i}e_{j-1,i}}
{\sqrt{
\bigl(\sum_{j\in\mathcal{W}_k^-}e_{j,i}^2\bigr)
\bigl(\sum_{j\in\mathcal{W}_k^-}e_{j-1,i}^2\bigr)
+\varepsilon_a}},
\nonumber\\
c_{k,i}&=
\frac{\sum_{j\in\mathcal{W}_k}e_{j,i}\xi_{j,i}}
{\sqrt{
\bigl(\sum_{j\in\mathcal{W}_k}e_{j,i}^2\bigr)
\bigl(\sum_{j\in\mathcal{W}_k}\xi_{j,i}^2\bigr)
+\varepsilon_c}}.
\end{align}
Here, $q_{k,i}$ is the log mean-square normalized innovation, $\mu_{k,i}$ is the mean normalized innovation, $a_{k,i}$ is the normalized lag-one innovation correlation, and $c_{k,i}$ is the normalized correlation between innovations and preceding reference velocities. The positive constants $(\varepsilon_q,\varepsilon_a,\varepsilon_c) =(10^{-2},10^{-2},10^{-6})$ ensure numerical stability.

In addition to these reference-KF statistics, the network receives $m_{k-1}\in\mathbb{R}^6$, which encodes the deviation of its previous adaptive model from the corresponding time-invariant model used to initialize that variant. For the structured variant, $m_{k-1}$ represents deviations in the three velocity parameters and three process-noise intensities from the constant fit. For the flexible variant, it analogously represents deviations in the diagonal entries of the velocity-dynamics and process-covariance blocks from its constant fit. The reference KF remains fixed and is used only to compute the statistics $(q_k,\mu_k,a_k,c_k)$.

Hence, the complete network input is
\begin{align}
\phi_k &=
\begin{bmatrix}
q_k^\top&
\mu_k^\top&
a_k^\top&
c_k^\top&
m_{k-1}^\top
\end{bmatrix}^{\top}
\in\mathbb{R}^{18},
\label{eq:architecture_input}
\end{align}
where $q_k,\mu_k,a_k,c_k\in\mathbb{R}^{3}$ and $m_{k-1}\in\mathbb{R}^{6}$. \rev[R1\_2]{The first four feature groups summarize recent measurement-model mismatch observed by the fixed reference KF, while $m_{k-1}$ provides feedback on the adaptive model's previous parameters. The mapping $f_\theta$ is implemented as a multilayer perceptron (MLP), and the measurement update uses the jointly learned time-invariant covariance $R^{\mathrm{ML}}$.}

\rev[R1\_2]{Computationally, relative to a standard KF, HAKF additionally evaluates the fixed reference KF, computes the windowed statistics, and performs one MLP forward pass at each time step. Computing the windowed statistics requires $\mathcal{O}(W)$ operations per axis, while the network architecture is fixed. Thus, for fixed state dimension, $W$, and network architecture, the per-step computational overhead is fixed and the total filtering cost scales linearly with the sequence length.}

\subsection{Cost Function}
\label{sec:objective}
Given the hybrid filtering architecture described above, we consider the marginal maximum likelihood estimation objective. Under the correct model specification and the KF assumptions, the predictive measurement distribution is
\begin{align}
\label{pred_meas}
z_k \mid z_{1:k-1}
\sim
\mathcal{N}(H_k \hat{x}_{k|k-1}, S_k).
\end{align}
Substituting this predictive distribution into \eqref{marginal_likelihood} yields a closed-form expression for the marginal likelihood. Maximizing this likelihood is equivalent to minimizing the innovation negative log-likelihood (NLL), which gives the cost function:
\begin{align}
\label{cost_function}
\mathcal{L}(\theta)
= \frac{1}{2}\sum_{k=1}^{T} \left(\nu_k^\top S_k^{-1} \nu_k + \log \det S_k\right),
\end{align}
where the additive constant independent of $\theta$ has been omitted. \rev[R1\_2]{Each term evaluates the one-step-ahead prediction of the next measurement, so HAKF can be trained in a self-supervised manner without ground-truth states.}

Under the assumption that the innovations are independent and Gaussian, minimizing the cost function in \eqref{cost_function} yields first-order optimality conditions that include a weighted covariance-matching term on the innovation sequence. This connection has been formally established in the data assimilation literature \cite{chapnik_properties_2004,tandeo_review_2020}. Under model misspecification, however, the resulting Kalman gain is generally suboptimal, and the innovation sequence becomes temporally correlated \cite{zhang_identification_2020}, violating the independence assumptions implicit in \eqref{cost_function}.   \rev[R2\_3]{In this regime, \eqref{cost_function} can be interpreted as a quasi-likelihood objective. Classical quasi-maximum likelihood estimation (QMLE) theory establishes convergence to a pseudo-true parameter only under additional assumptions, including regularity and identifiability \cite{white_maximum_1982,gallant_unified_1988}. We do not establish these conditions for the closed-loop neural adaptation in HAKF. QMLE is used here as statistical motivation for the innovation NLL rather than as a convergence guarantee for the learned model. We also do not claim that gradient-based optimization reaches a global minimizer of this non-convex objective.} A separate limitation of the innovation NLL is that it does not include an explicit penalty on cross-lag innovation correlations, so temporal dependence may persist after training.

To make this temporal structure available to the adaptive model, the reference-KF features include statistics of recent normalized innovations together with lag-one innovation and innovation--velocity correlations defined in \eqref{eq:architecture_correlations}. In this way, HAKF integrates principles from likelihood-based, covariance-matching, and correlation-based classical adaptive Kalman filtering methods through its cost function and architecture.

\rev[AE, R1\_2]{Training proceeds in two stages. First, the time-invariant counterpart of each variant is fitted together with $R$ by minimizing \eqref{cost_function} on the training data, using constant $(\Gamma,\Sigma)$ for the structured variant and a constant velocity-transition block and $Q$ for the flexible variant. The network is then initialized at the corresponding constant model, $R^{\mathrm{ML}}$ is initialized at its fitted value, and $\theta$ and $R^{\mathrm{ML}}$ are optimized jointly using Adam~\cite{kingma_adam_2017}. For each measurement sequence, the initial estimate $(\hat{x}_0,P_0)$ is obtained by weighted least squares from the first two measurements using $F^{\mathrm{CV}}$ and $R^{\mathrm{ML}}$~\cite{simon_optimal_2006}.}

\subsection{Classification via Generalized Bayesian Inference}
\label{sec:GBI}
The KF can be interpreted as a generative model that induces a predictive measurement likelihood through the distribution defined in \eqref{pred_meas}. Model classification can therefore be performed by computing the posterior over candidate models using Bayes' rule. 

Under model misspecification, however, direct likelihood-based calculation may lead to overconfident or inconsistent posterior estimates \cite{knoblauch_generalized_2019}. To mitigate this issue, we adopt generalized Bayesian inference (GBI) using a tempered likelihood \cite{holmes_assigning_2017-1},
\begin{align}
p_\alpha(M_j \mid Z)\propto p(Z \mid M_j)^{\alpha} \; p(M_j),
\end{align}
where $J$ is the number of candidate models, and a single $\alpha\in(0,1]$ is shared across all models to control the influence of the likelihood on the posterior.

Various approaches for selecting the tempering parameter have been proposed in the literature \cite{wu_comparison_2023}. \rev[AE]{In this work, $\alpha$ is selected by grid search to minimize the multiclass Brier score on a validation dataset,}
\begin{align}
\alpha^*
=
\arg\min_{\alpha\in\mathcal{A}}
\frac{1}{N_v}
\sum_{i=1}^{N_v}
\sum_{j=1}^{J}
\left(
p_\alpha(M_j\mid Z^{(i)})
-y_{i,j}
\right)^2,
\end{align}
where $N_v$ is the number of validation segments, $Z^{(i)}$ denotes the measurement sequence of the $i$-th validation segment, $y_{i,j}\in\{0,1\}$ is the corresponding class indicator, and $\mathcal{A}$ denotes the temperature-search grid. When a uniform prior is used, applying the same positive $\alpha$ to all candidate models rescales the posterior probabilities without changing the maximum-likelihood class decision.
\section{Experimental Setup}
\label{sec:experimental_setup}
\subsection{Datasets}
\label{sec:datasets}
\begin{figure}[t]
    \centering
    \subfloat[]{%
        \includegraphics[width=0.48\linewidth]{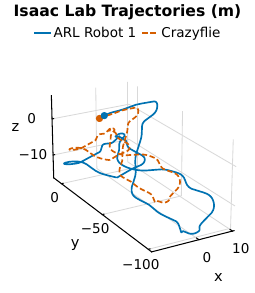}
        \label{fig:traj_a}
    }
    \hfill
    \subfloat[]{%
        \includegraphics[width=0.48\linewidth]{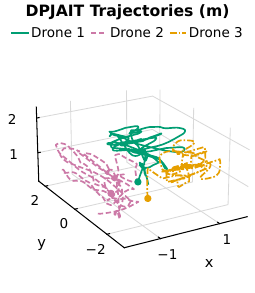}
        \label{fig:traj_b}
    }
    \caption{Example trajectories used in the experiments. (a) Isaac Lab flights of the ARL Robot~1 and Crazyflie. (b) Real-world trajectories from the DPJAIT dataset.}
    \label{fig:traj}
\end{figure}

\subsubsection{Isaac Lab Simulation}
Simulated flights of two quadrotor platforms, the ARL Robot~1 and Crazyflie, are generated in Isaac Lab~3.0~\cite{nvidia_isaac_2025} (Fig.~\ref{fig:traj}(a)). Each flight is 60~s long and is recorded at 20~Hz. Velocity commands composed of multisine signals with varying excitation levels and cross-axis coupling are used to intentionally generate motion that departs from the nominal CV model. \rev[AE]{For each drone, the number of training flights is varied as $N\in\{4,8,16,32,64,128,256\}$. Additionally, 512 and 400 disjoint flights are used for validation and testing, respectively.}

\subsubsection{DPJAIT}
The DPJAIT dataset~\cite{rosner_multimodal_2025} contains motion-capture recordings of indoor flights from three quadrotor platforms (Fig.~\ref{fig:traj}(b)). The drone position is defined as the centroid of its tracked markers. The original 100-Hz trajectories are subsampled every fifth frame, yielding the same sampling interval $\Delta t=0.05$~s used in the simulated experiments. Velocity ground truth is computed from the original position trajectories before subsampling. Only the airborne portions of each trajectory are retained, and tracks consisting entirely of hovering are excluded. For each drone, two tracks are held out for testing, one for validation, and the remaining tracks provide 36, 194, and 70 ten-second training windows, corresponding to approximately 6, 33, and 12~min of flight, respectively.

\subsubsection{Measurements}
Position-only measurements are generated by adding zero-mean white Gaussian noise with a standard deviation of 10 cm along each axis. The measurement-noise covariance used to generate the observations is not provided to the filters. Both HAKF and KF baselines estimate the covariance from the training data. For DPJAIT, each test track is evaluated over 50 independent measurement-noise realizations.

\subsection{HAKF Training}
\label{sec:training}
The mapping $f_\theta$ is implemented as an MLP with two fully connected layers of 128 and 256 hidden units, and the feature-window length is set to $W=10$. The network is trained on ten-second windows using Adam with a learning rate of $10^{-3}$ and early stopping on the validation NLL. \rev[AE]{For each drone and training-set size, five networks of each variant are trained with different random seeds, and the reported results are averaged over them. The interpolation parameter $\lambda$ of the flexible variant is selected from $\{0.1,0.2,0.5,1\}$ by the validation NLL.}

\subsection{Baselines}
\label{sec:baselines}
\subsubsection{Tracking}
The proposed HAKF is compared with a classical CV KF and a supervised recurrent neural network (RNN). The KF baseline uses $F^{\mathrm{CV}}$ together with a white-noise-acceleration process covariance $Q^{\mathrm{ref}}$ with per-axis intensity $q_i^{\mathrm{ref}}$~\cite{baisa_derivation_2020-1} and a diagonal measurement covariance $R^{\mathrm{ref}}=\operatorname{diag}(r_x^{\mathrm{ref}},r_y^{\mathrm{ref}},r_z^{\mathrm{ref}})$. \rev[AE, R1\_4]{These parameters are fitted for each drone by minimizing \eqref{cost_function} on the same training data used by HAKF.} This filter also serves as the fixed reference KF used to construct the HAKF input features.

The learning-based baseline uses a gated recurrent unit (GRU)~\cite{cho2014learningphraserepresentationsusing} to map the most recent 20 measurements (1~s) to the current position and velocity. The inputs are represented using measurement increments and offsets relative to the current measurement. The network is trained using ground-truth states by minimizing the squared state error. The hidden dimension and number of recurrent layers are selected from $\{32,64,128\}$ and $\{1,2\}$, respectively, using validation state RMSE, and five networks are trained with different random seeds for each drone. Unlike the KF-based methods, the RNN produces point estimates only and therefore does not provide covariance estimates for consistency evaluation.

\subsubsection{Classification}
Three learning-based classification baselines are considered. The Random Forest (RF)~\cite{breiman_random_2001} classifies each measurement segment using summary statistics of velocity and acceleration obtained by differencing the measurements. The RF + KF baseline computes the same statistics after filtering the measurements with a single CV KF whose noise parameters are fitted on the pooled training data. For both RF baselines, the number of trees and depth are selected using the validation Brier score.

The RNN classifier uses a bidirectional GRU that processes the velocity and acceleration obtained by differencing the measurements over the full segment. Its hidden dimension, number of layers, and number of training epochs are selected using the validation Brier score, with one classifier trained for each segment length (10, 20, and 40~s). The inputs are averaged over 0.25-s blocks to reduce measurement noise. 

\rev[R1\_4]{The RF and RNN classifiers are trained using labeled segments from all classes, whereas each HAKF model is trained only on the unlabeled measurement sequences of its own drone.} Five RF and RNN classifiers are trained with different random seeds.

\subsection{Evaluation Metrics}
\subsubsection{Tracking}
\label{sec:estimation_metric}
Estimation accuracy is evaluated using full state, position, and velocity RMSE. Predictive performance is measured using the innovation NLL corresponding to \eqref{cost_function}, averaged over the evaluated time steps. Filter consistency is evaluated using the average NEES (ANEES) and average NIS (ANIS), whose expected values for a consistent filter are the state and measurement dimensions, 6 and 3, respectively. The results are averaged across the two Isaac Lab and three DPJAIT drone platforms.

\subsubsection{Classification}
\label{sec:classification_metric}
Classification is evaluated on non-overlapping segments of 10, 20, and 40~s. Each segment is processed independently by the HAKF filters and the trained classification baselines. Classification performance is measured using balanced accuracy, which is unchanged by the common tempering parameter $\alpha$ under the uniform class prior.

\section{Results}
\label{sec:results}
\begin{figure*}[t]
\centering
\includegraphics[width=\textwidth]{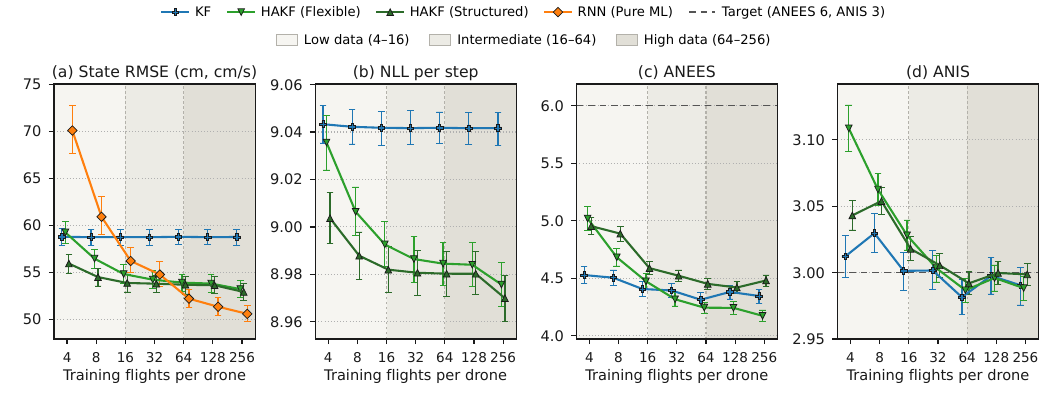}
\caption{Isaac Lab tracking performance versus the number of 60-s training flights per drone: (a) state RMSE over all six state components, (b) per-step NLL, (c) ANEES, and (d) ANIS. Markers show mean values and error bars indicate 95\% confidence intervals over the test flights. Dashed lines indicate the expected ANEES and ANIS values of 6 and 3, respectively.}
\label{fig:isaaclab_tracking}
\end{figure*}

\begin{table}[t]
\centering
\caption{Tracking performance on DPJAIT with 10-cm position-measurement noise, averaged across three drones and 50 noise realizations.}
\label{tab:estimation_dpjait}
\scriptsize
\setlength{\tabcolsep}{3pt}
\begin{tabular}{@{}lccccc@{}}
\toprule
Method & Pos. RMSE $\downarrow$ & Vel. RMSE $\downarrow$ & NLL $\downarrow$ & ANEES & ANIS \\
\midrule
KF (Fixed $F,Q,R$)$^*$& 8.83 & 37.37 & 8.86 & 9.10 & \textbf{3.08}\\
HAKF (Flexible)$^*$ & 8.43 & 35.18 & 8.82 & 15.55 & 3.12 \\
HAKF (Structured)$^*$ & \textbf{8.30} & \textbf{33.81} & \textbf{8.81} & \textbf{6.60} & \textbf{3.08} \\
RNN (Pure ML)$^\dagger$ & 8.68 & 34.66 & --- & --- & --- \\
\midrule
Target & --- & --- & --- & 6 & 3 \\
\bottomrule
\end{tabular}

\footnotesize{
$^*$ Fitted (KF) or trained (HAKF) by minimizing \eqref{cost_function}. $^\dagger$ Trained using the ground-truth states with squared state error. Position and velocity RMSE are in cm and cm/s, respectively. NLL is reported per step.
}
\end{table}

\begin{figure*}[t]
\centering
\includegraphics[width=\textwidth]{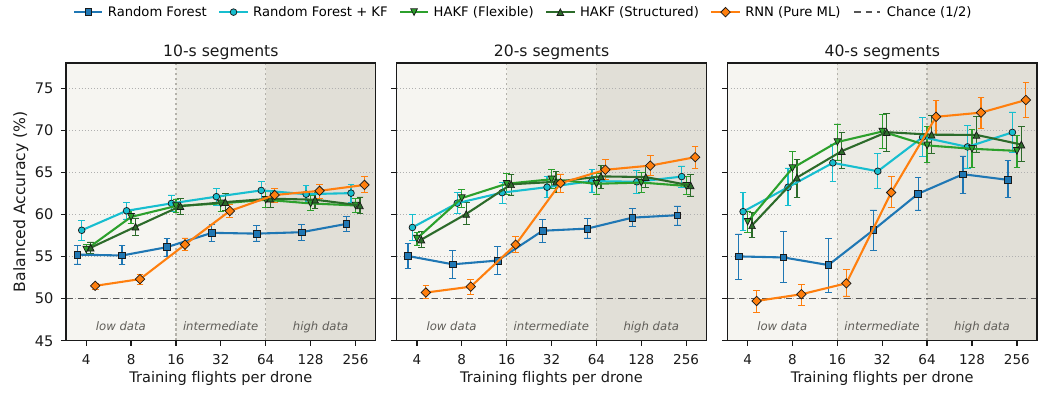}
\caption{Balanced classification accuracy on Isaac Lab for 10-, 20-, and 40-s segments versus the number of 60-s training flights per drone, evaluated on 400 test flights per drone. Markers show mean accuracy and error bars indicate 95\% confidence intervals over the test flights. Shaded regions mark the low, intermediate, and high-data regimes (4--16, 16--64, and 64--256 flights).}
\label{fig:isaaclab_classification}
\end{figure*}

\begin{figure}[t]
\centering
\includegraphics[width=\columnwidth]{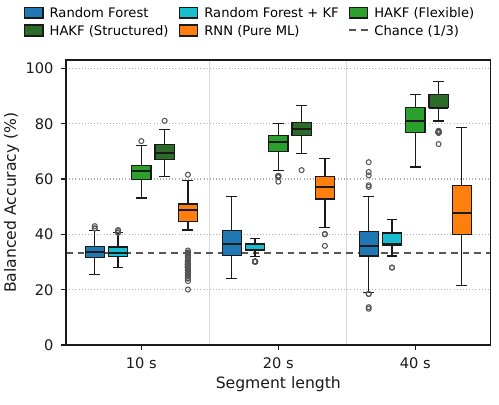}
\caption{Balanced classification accuracy on the DPJAIT dataset with simulated 10-cm position-measurement noise for 10-, 20-, and 40-s segments. Box plots summarize the distributions across five trained classifiers and 50 measurement-noise realizations. Whiskers extend to 1.5 times the interquartile range with outliers shown individually.}
\label{fig:dpjait_classification}
\end{figure}

\subsection{Tracking Performance}
Fig.~\ref{fig:isaaclab_tracking} shows the tracking performance on Isaac Lab as the number of training flights increases. HAKF (Structured) achieves lower state RMSE than the KF baseline across all training-set sizes, while HAKF (Flexible) improves upon the KF from eight training flights onward. The gap between the two HAKF variants is largest in the low-data regime and narrows as the training-set size increases. This suggests that the physically structured parameterization provides a useful inductive bias when data are limited, whereas the flexible variant requires more data to approach its performance. The supervised RNN performs substantially worse in the low-data regime, but improves with additional training data and achieves the lowest state RMSE in the high-data regime.

Both HAKF variants also achieve lower innovation NLL than the KF across all training-set sizes, following a similar trend to the RMSE. The ANEES values of the KF and HAKF variants remain below the expected value, indicating conservative state covariance estimates, while the ANIS values remain close to the expected value.

Table~\ref{tab:estimation_dpjait} summarizes the tracking performance on the DPJAIT test trajectories. HAKF (Structured) achieves the best overall performance, reducing both position and velocity RMSE relative to the KF baseline while maintaining an ANEES close to the expected value of 6. Its innovation NLL is comparable to that of HAKF (Flexible). The RNN, although trained using the ground-truth states, improves on the KF primarily in velocity RMSE and remains less accurate than HAKF (Structured).

HAKF (Flexible) also improves estimation accuracy and NLL relative to the KF, but its ANEES of 15.55 indicates overconfident state covariance estimates. \rev[R2\_1]{Proposition~\ref{thm:full_q_nonidentifiable} in Section~\ref{sec:identifiability_cv} provides one possible explanation for this result. Even for the time-invariant CV model with known dynamics, an unrestricted process covariance is not identifiable from position measurements. These results do not establish identifiability of the time-varying neural-network adaptation but show how flexible joint adaptation of the dynamics and process covariance can leave the latent state uncertainty weakly constrained by the innovation likelihood. Consistent with this interpretation, the excess ANEES is concentrated in the velocity components, with velocity and position-only NEES values of 12.2 and 3.3, respectively, against an expected value of 3 for each.} 

The consistency results also reflect model mismatch beyond identifiability. On Isaac Lab, the ANEES values remain below the expected value for all filters, whereas on DPJAIT, they exceed 6. The measurement noise is white by construction in both datasets, but residual dynamics mismatch not captured by the learned transition must still be represented through the assumed white process noise. As a result, neither parameterization is expected to achieve exact statistical consistency when this assumption is violated.

\subsection{Classification Performance}
Fig.~\ref{fig:isaaclab_classification} shows the balanced classification accuracy on Isaac Lab for 10-, 20-, and 40-s measurement segments. In the low-data regime, both HAKF variants outperform the RNN across all segment lengths. The RF + KF baseline is the strongest competing method and performs comparably to HAKF.  As the training set grows, the accuracy of HAKF plateaus, whereas the supervised RNN continues to improve and achieves the highest accuracy at the largest training-set sizes. This behavior highlights the different data requirements of the approaches. HAKF benefits from the model-based structure when training data are limited, whereas the RNN benefits more strongly from larger labeled datasets. 

Fig.~\ref{fig:dpjait_classification} shows the balanced classification accuracy on the DPJAIT test trajectories. HAKF (Structured) achieves the highest median accuracy at all segment lengths, followed by HAKF (Flexible). The RF-based baselines remain near chance level, while the RNN improves upon them but remains below both HAKF variants. With only a few training tracks per drone, the motion statistics used by the RF-based classifiers may not generalize well across trajectories.

For both HAKF variants, classification accuracy increases with the segment length, consistent with the accumulation of predictive evidence over longer measurement sequences. These results are particularly notable because each HAKF model is trained only on its own drone's measurements, whereas the classification baselines are trained directly on labeled segments.

\section{Conclusion and Future Work}
\label{sec:conclusion}
This article proposed a self-supervised HAKF framework that integrates model-based filtering with neural network-based adaptation using only measurement data. To determine which adaptive parameterizations can be learned from measurements alone, we analyzed the structural identifiability of the flexible and structured CV parameterizations. For the time-invariant CV model, an unrestricted process covariance is not identifiable from position measurements, whereas coupling the dynamics and process covariance through the structured CV parameterization removes this ambiguity under the stated assumptions. HAKF adapts these parameterizations over time while preserving the probabilistic filtering structure, allowing the same learned models to perform both state estimation and classification through GBI.

Experimental results demonstrate the data efficiency of the proposed framework. On Isaac Lab, HAKF (Structured), which restricts the adaptation to a physically structured model family whose time-invariant counterpart is identifiable, improves upon the KF baseline across all training-set sizes and shows the largest advantage over the flexible variant and supervised RNN when training data are limited. As more data become available, the performance gap between the HAKF variants narrows and the supervised RNN becomes competitive and, with the largest training sets, more accurate. On DPJAIT, HAKF (Structured) achieves the best overall tracking performance and the highest classification accuracy among the compared methods while maintaining state covariance consistency close to the expected level. In contrast, HAKF (Flexible) achieves similar predictive likelihood but becomes overconfident, illustrating how flexible joint adaptation can leave latent-state uncertainty weakly constrained by the innovation likelihood.

\rev[R1\_3, R2\_1, R2\_2]{The identifiability analysis applies only to constant parameters in the time-invariant linear models and therefore does not establish identifiability of the time-varying neural-network adaptation. The stability of the resulting time-varying filter is also not analyzed.} Moreover, the observed data-efficiency gains are empirical, and non-identifiability allows but does not necessarily imply covariance miscalibration. HAKF also retains modeling assumptions inherited from the standard KF. In particular, residual dynamics not captured by the learned transition are still represented through white Gaussian process noise, which may not adequately describe temporally correlated model mismatch. In addition, the DPJAIT experiments use a limited number of recorded trajectories with simulated measurement noise. \rev[R2\_2, R2\_4]{Future work will analyze the stability of HAKF, extend the identifiability analysis to time-varying and nonlinear models, incorporate richer representations of residual dynamics, and use real sensor measurements where the process and measurement noises may be correlated. }
\appendices
\section{Proof of Structured CV Identifiability}
\label{app:structured_identifiability}
\rev[R1\_3, R2\_1]{
We prove Theorem~\ref{thm:structured_cv_identifiability} for one position--velocity axis. Denote the constant parameters by $(\gamma,\sigma^2,r)$. With sampling interval $\Delta t>0$, define
\begin{align}
\rho &= e^{-\gamma\Delta t}>0, &
\psi &=\int_0^{\Delta t} e^{-\gamma s}\,ds>0.
\label{eq:appendix_rho_psi}
\end{align}
Writing $\dot p_k$ for the scalar velocity state, the exact discrete-time dynamics and measurement equation are
\begin{align}
p_{k+1}&=p_k+\psi \dot p_k+w_k^p,\nonumber\\
\dot p_{k+1}&=\rho \dot p_k+w_k^v,\qquad z_k=p_k+v_k,
\label{eq:appendix_1d_discrete}
\end{align}
where $w_k=[w_k^p,w_k^v]^\top\sim\mathcal{N}(0,Q(\gamma,\sigma^2))$ and $v_k\sim\mathcal{N}(0,r)$ are mutually independent white process and measurement noises, respectively.
The initial Gaussian state is independent of both noise sequences. The process covariance factorizes as
\begin{align}
Q(\gamma,\sigma^2)
&=\sigma^2\bar Q(\gamma),\qquad
\bar Q(\gamma)=
\begin{bmatrix}\bar q_{pp}&\bar q_{pv}\\\bar q_{pv}&\bar q_{vv}\end{bmatrix},\nonumber\\
\bar q_{pp}&=\int_0^{\Delta t}\!\psi(s)^2\,ds,\quad
\bar q_{pv}=\int_0^{\Delta t}\!\psi(s)e^{-\gamma s}\,ds,\nonumber\\
\bar q_{vv}&=\int_0^{\Delta t}\!e^{-2\gamma s}\,ds,\quad
\psi(s)=\int_0^s\!e^{-\gamma\tau}\,d\tau.
\label{eq:appendix_q_elements}
\end{align}
For every finite real $\gamma$, $\psi(s)>0$ for $s>0$; therefore, $\bar q_{pp}>0$ and $\bar q_{pv}>0$. Furthermore, $\bar Q\succeq0$ by construction.

\emph{Step 1: Identify the velocity parameter.} Define
\begin{align}
d_k=z_{k+1}-z_k = \psi \dot p_k+w_k^p+v_{k+1}-v_k.
\label{eq:appendix_dk}
\end{align}
For $j\ge2$, the process and measurement noise terms associated with $d_k$ and $d_{k+j}$ occur over nonoverlapping time intervals and are independent. Hence, expressing $\dot p_{k+j}$ through $\dot p_{k+1}$ gives
\begin{align}
C_j(k)&\triangleq\operatorname{Cov}(d_k,d_{k+j})=\psi\rho^{j-1}\operatorname{Cov}(d_k,\dot p_{k+1})\nonumber\\
&=\psi\rho^{j-1}\bigl[\psi\rho\operatorname{Var}(\dot p_k)
+\sigma^2\bar q_{pv}\bigr].
\label{eq:appendix_increment_cov}
\end{align}
Because $\psi,\rho,\sigma^2$, and $\bar q_{pv}$ are strictly positive and $\operatorname{Var}(\dot p_k)\ge0$, we have $C_2(k)>0$. Consequently $C_3(k)/C_2(k)=\rho$, which uniquely determines $\gamma=-\log(\rho)/\Delta t$.

\emph{Step 2: Identify the measurement noise variance and process noise intensity.}
Once $\rho$ is known, define $u_k=d_k-\rho d_{k-1}$ for $k\ge1$. Using $\dot p_k=\rho \dot p_{k-1}+w_{k-1}^v$ in \eqref{eq:appendix_dk}, the unobserved velocity cancels:
\begin{align}
u_k &= w_k^p+\psi w_{k-1}^v-\rho w_{k-1}^p +v_{k+1}-(1+\rho)v_k +\rho v_{k-1}.
\label{eq:appendix_uk}
\end{align}
The covariance $\Lambda_j=\operatorname{Cov}(u_k,u_{k+j})$ is now independent of $k$ and depends only on a finite number of white-noise terms. At lag two, the only shared noise term is $v_{k+1}$, with coefficients $1$ and $\rho$, respectively. Hence
\begin{align}
\Lambda_2=\rho r.
\label{eq:appendix_L2}
\end{align}
At lag zero, independence between process and measurement noises, and between different sampling intervals, yields
\begin{align}
\Lambda_0&=\sigma^2\ell_0(\gamma)+\kappa_0(\rho)r,\nonumber\\
\ell_0(\gamma)&=(1+\rho^2)\bar q_{pp}
-2\rho\psi\bar q_{pv}+\psi^2\bar q_{vv},\nonumber\\
\kappa_0(\rho)&=1+(1+\rho)^2+\rho^2.
\label{eq:appendix_L0}
\end{align}
The process coefficient can equivalently be written as
\begin{align}
\ell_0(\gamma)=\bar q_{pp}+
\begin{bmatrix}-\rho&\psi\end{bmatrix}
\bar Q(\gamma)
\begin{bmatrix}-\rho\\\psi\end{bmatrix}>0,
\label{eq:appendix_l0_positive}
\end{align}
since $\bar q_{pp}>0$ and $\bar Q\succeq0$. Thus, $r$ and $\sigma^2$ are uniquely recovered as
\begin{align}
r &= \frac{\Lambda_2}{\rho},\qquad 
\sigma^2 = \frac{\Lambda_0-\kappa_0(\rho)r}{\ell_0(\gamma)}.
\label{eq:appendix_sigma_recovery}
\end{align}
This establishes identifiability for the one-dimensional model.

\emph{Three-axis extension.} With diagonal $\Gamma$, $\Sigma$, and $R$,
the dynamics and noise processes decouple by axis. Applying the one-dimensional proof to the measurement sequence of each axis establishes identifiability of $\Gamma$, $\Sigma$, and $R$. This completes the proof of Theorem~\ref{thm:structured_cv_identifiability}.}
\section{Likelihood Geometry of the CV Parameterizations}
\label{app:likelihood_geometry}
\begin{figure}[t]
  \centering
  \includegraphics[width=\linewidth]{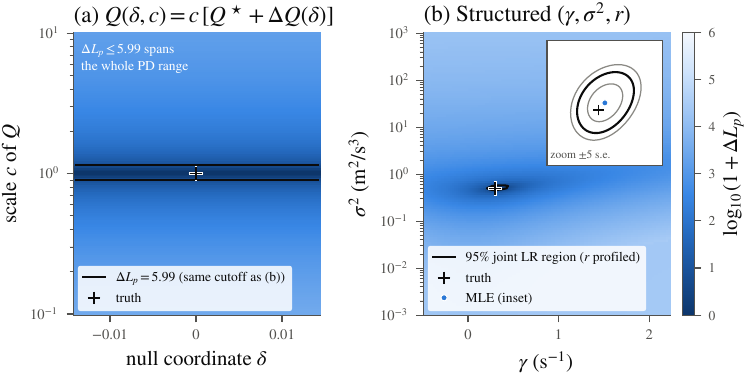}
  \caption{Profile-likelihood geometry of the CV parameterizations. (a) Unrestricted $Q$ exhibits a flat ridge along the null coordinate $\delta$; the black contour denotes the reference level $\Delta L_p=5.99$. (b) The structured model exhibits a localized optimum, with the black contour denoting the approximate 95\% joint likelihood-ratio region.}
  \label{fig:identifiability_likelihood}
\end{figure}
\rev[AE, R2\_5]{
Figure~\ref{fig:identifiability_likelihood} compares the profile-likelihood geometry of the unrestricted and structured CV parameterizations using simulated position measurements.
For panel (a), let $\vartheta=(\delta,c)$. For panel (b), let $\vartheta=(\gamma,\sigma^2)$. With measurement-noise variance $r$ profiled out, define $\ell_p(\vartheta)=\max_{r>0}\log p(z_{1:T}\mid\vartheta,r)$. The likelihood-ratio loss is 
\begin{align}
  \Delta L_p(\vartheta) =2\left[\max_{\vartheta'}\ell_p(\vartheta')-\ell_p(\vartheta)\right].  
\end{align}

For the unrestricted model, panel (a) fixes the transition at its true value $F^\star$ and varies the process covariance
\begin{equation}
Q(\delta,c)=c\left[Q^\star+\Delta Q(\delta)\right],\quad
\Delta Q(\delta)=E_\delta-F^\star E_\delta F^{\star\top},
\label{eq:appendix_null_slice}
\end{equation}
over the positive-definite domain, with $c>0$ and $E_\delta=\operatorname{diag}(0,\delta)$, where $F^\star$, $Q^\star$, and $P_0^\star$ denote the transition matrix and the process and initial-state covariances used to simulate the measurements. The initial covariance is shifted accordingly to $P_0^\star+c\operatorname{diag}(0,\delta)$, which leaves the measurement-sequence distribution unchanged along $\delta$, as established in Proposition~\ref{thm:full_q_nonidentifiable}.

For the structured model, $F$ and $Q$ are coupled through $(\gamma,\sigma^2)$. Panel (b) shows a localized profile-likelihood optimum and bounded 95\% joint likelihood-ratio region after profiling over $r$. This finite-sample illustration agrees with the structural identifiability result of Theorem~\ref{thm:structured_cv_identifiability}.}
\bibliographystyle{IEEEtran}
\bibliography{HAKF}

\begin{IEEEbiography}
[{\includegraphics[width=1in,height=1.25in,clip]{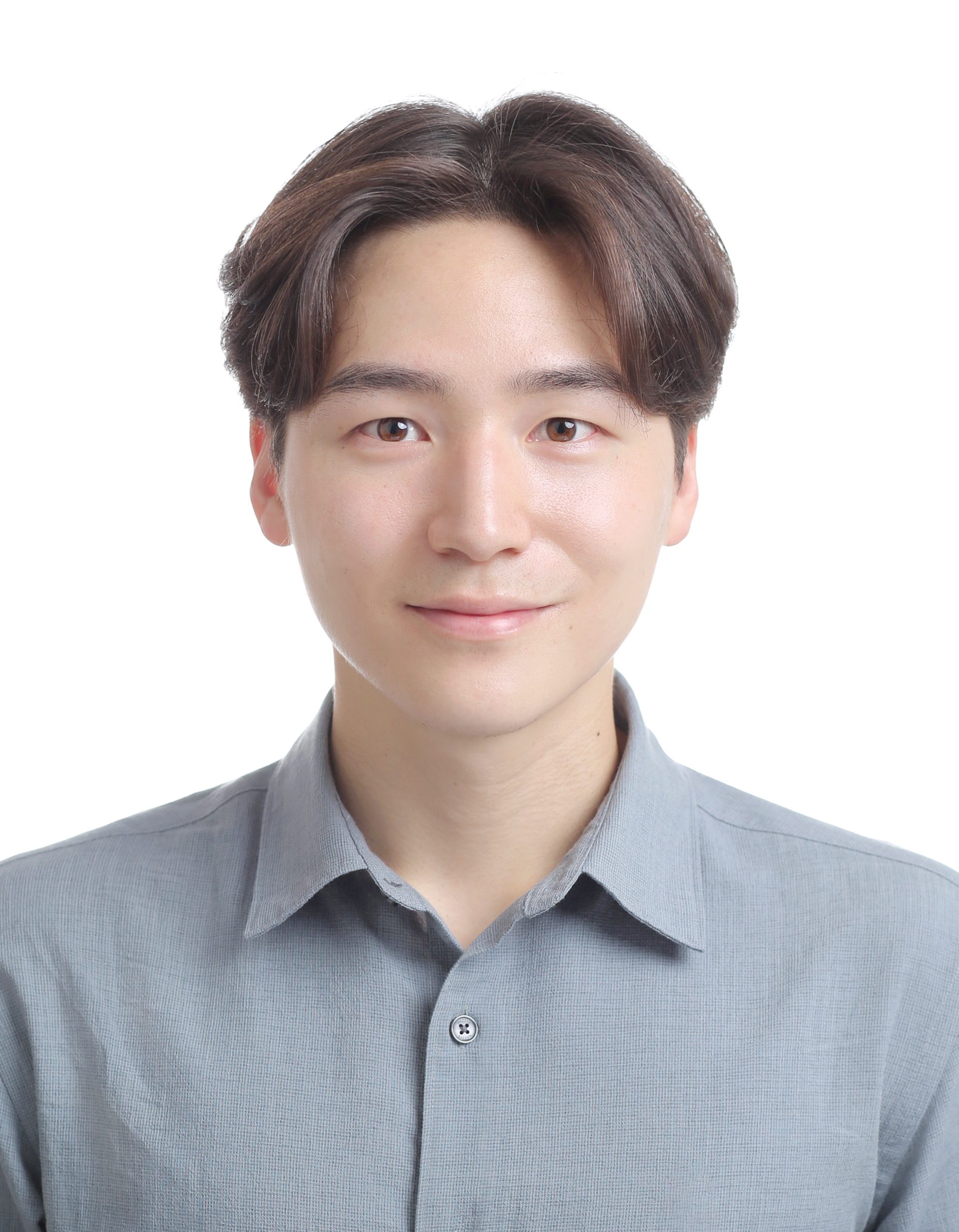}}]{Jiho Lee}
(Student Member, IEEE) received the B.S. degree in economics and the B.A.S. degree in computer science from the University of Pennsylvania, Philadelphia, PA, USA, in 2020. He is currently pursuing the Ph.D. degree in aerospace engineering at the University of Colorado Boulder, Boulder, CO, USA. His research interests include probabilistic inference, decision-making under uncertainty, and machine learning.
\end{IEEEbiography}
\begin{IEEEbiography}
[{\includegraphics[width=1in,height=1in,clip]{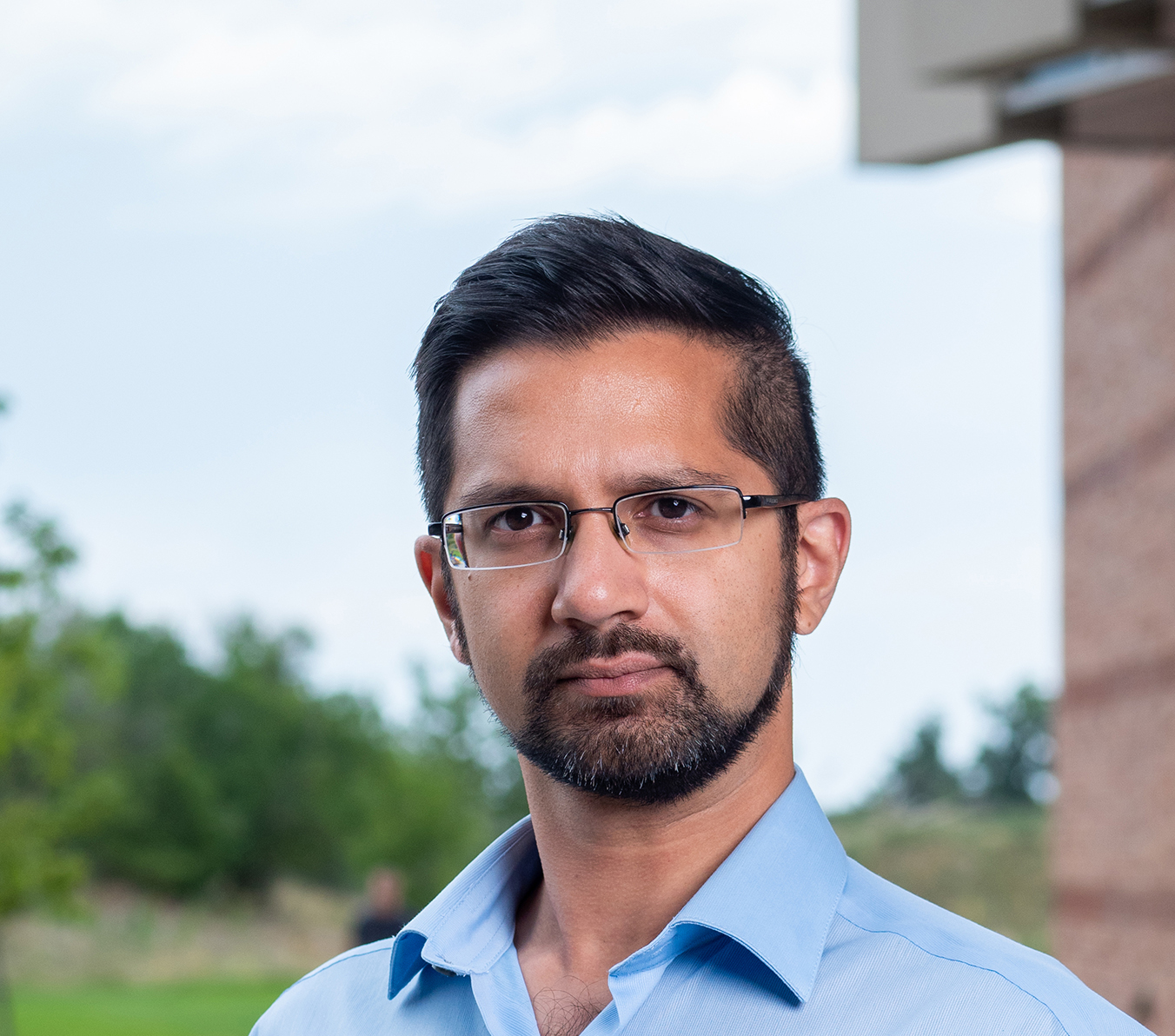}}]{Nisar Ahmed} (Member, IEEE) received the B.S. degree in engineering from Cooper Union, New York, NY, USA, in 2006, and the Ph.D. degree in mechanical engineering from Cornell University, Ithaca, NY, USA, in 2012. He is currently an Associate Professor and H.J. Smead Faculty Fellow with the Smead Aerospace Engineering Sciences Department, University of Colorado Boulder, Boulder, CO, USA. He is the Director with the Cooperative Human-Robot Intelligence (COHRINT) Lab. His research interests include probabilistic modeling, estimation and control of autonomous systems, human-robot/machine interaction, sensor fusion, and decision-making under uncertainty.
\end{IEEEbiography}

\begin{IEEEbiography}
[{\includegraphics[width=1in,height=1.25in,clip]{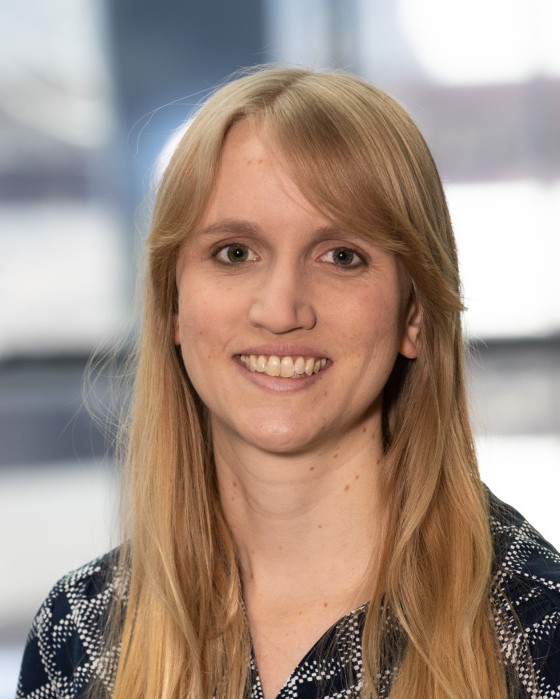}}]{Rebecca Russell} received the B.S. degree in physics from the California Institute of Technology, Pasadena, CA, USA, in 2010 and the Ph.D. degree in physics from the Massachusetts Institute of Technology, Cambridge, MA, USA, in 2016.

Since 2016, she has been a machine learning scientist at the Charles Stark Draper Laboratory, Inc., where she is now a Principal Member of the Technical Staff. Her research focuses on uncertainty-aware deep learning for autonomy and aerospace applications.
\end{IEEEbiography}

\end{document}